\documentclass[runningheads]{llncs}
\usepackage{amsmath}
\usepackage{graphicx}
\usepackage{booktabs}
\usepackage{subcaption} 
\usepackage{amsfonts}
\usepackage{hyperref}
\usepackage{enumitem}
\usepackage{multirow} 
\usepackage{makecell}
\usepackage{pifont}
\usepackage{xcolor}
\usepackage[table]{xcolor}

\usepackage{marvosym}

\begin{document}
%
%\title{Contribution Title\thanks{Supported by organization x.}}
\title{Simple Lines, Big Ideas: Towards Interpretable Assessment of Human Creativity from Drawings}
\titlerunning{Towards Interpretable Creativity Assessment from Drawings}

%-----------------------------------
\author{Zihao Lin\inst{1} \and
Zhenshan Shi\inst{1} \and 
Sasa Zhao\inst{1} \and 
Hanwei Zhu\inst{3} \and
Lingyu Zhu\inst{2} \and
Baoliang Chen\inst{1}\textsuperscript{(\Letter)} \and 
Lei Mo\inst{1}\textsuperscript{(\Letter)} }

\institute{Department of Computer Science, South China Normal University, Guang Zhou, China\\
\email{blchen6-c@my.cityu.edu.hk, molei@scnu.edu.cn}\\
\and
Department of Computer Science, City University of Hong Kong, Hong Kong SAR, China
\and
College of Computing and Data Science, Nanyang Technological University, Singapore\\
}

\authorrunning{Z. Lin et al.}

\maketitle              % typeset the header of the contribution
\renewcommand{\thefootnote}{\fnsymbol{footnote}} 
\footnotetext[0]{\textsuperscript{(\Letter)} Corresponding author.} 

\begin{abstract}
Assessing human creativity through visual outputs, such as drawings, plays a critical role in fields including psychology, education, and cognitive science. However, current assessment practices still rely heavily on expert-based subjective scoring, which is both labor-intensive and inherently subjective.
In this paper, we propose a data-driven framework for automatic and interpretable creativity assessment from drawings. Motivated by the cognitive evidence proposed in \cite{Pencils_to_Pixels} that creativity can emerge from both  \textbf{\textit{what}} is drawn (content) and \textbf{\textit{how}} it is drawn (style), we reinterpret the creativity score as a function of these two complementary dimensions. Specifically, we first augment an existing creativity-labeled dataset with additional annotations targeting content categories. Based on the enriched dataset, we further propose a conditional model predicting content, style, and ratings simultaneously.
In particular, the conditional learning mechanism that enables the model to adapt its visual feature extraction by dynamically tuning it to creativity-relevant signals conditioned on the drawing’s stylistic and semantic cues.
Experimental results demonstrate that our model achieves state-of-the-art performance compared to existing regression-based approaches and offers interpretable visualizations that align well with human judgments. The code and annotations will be made publicly available at  \url{https://github.com/WonderOfU9/CSCA_PRCV_2025}

\end{abstract}

\section{Introduction}

Creativity is widely recognized as a critical skill in the modern educational and cognitive sciences landscape~\cite{unesco_creativity,p21_skills,runco_def}. For instance, UNESCO emphasizes creativity’s role in preparing learners for an uncertain future~\cite{unesco_creativity}, and creativity is explicitly listed as one of the core 21st-century skills for students~\cite{p21_skills}. Theoretical frameworks also highlight creativity’s dual criteria of novelty and appropriateness~\cite{runco_def,sternberg_creativity}, underscoring its importance in innovation and problem-solving. Assessing creativity reliably is therefore of great interest, both to identify individual strengths and to foster creative development in educational settings~\cite{p21_skills,runco_def}.

A common approach to measuring creative potential involves structured drawing or figural tasks. Classic tests such as the Torrance Test of Creative Thinking (TTCT) include a figural component where participants complete or embellish simple drawings~\cite{ttct}. Similar figure-drawing tasks (e.g., the Wallach–Kogan test) are routinely used in creativity research to elicit original visual ideas~\cite{wallach_kogan,plucker_review}. These tasks typically elicit sketches that are later scored by experts on dimensions like originality, elaboration, and unusualness~\cite{plucker_review,baer_scoring}. Such drawing-based paradigms are valued because they tap visual ideation processes that verbal tests cannot easily capture~\cite{plucker_review,baer_scoring}.

\begin{figure}[t]
  \centering
  \includegraphics[width=1.0\textwidth]{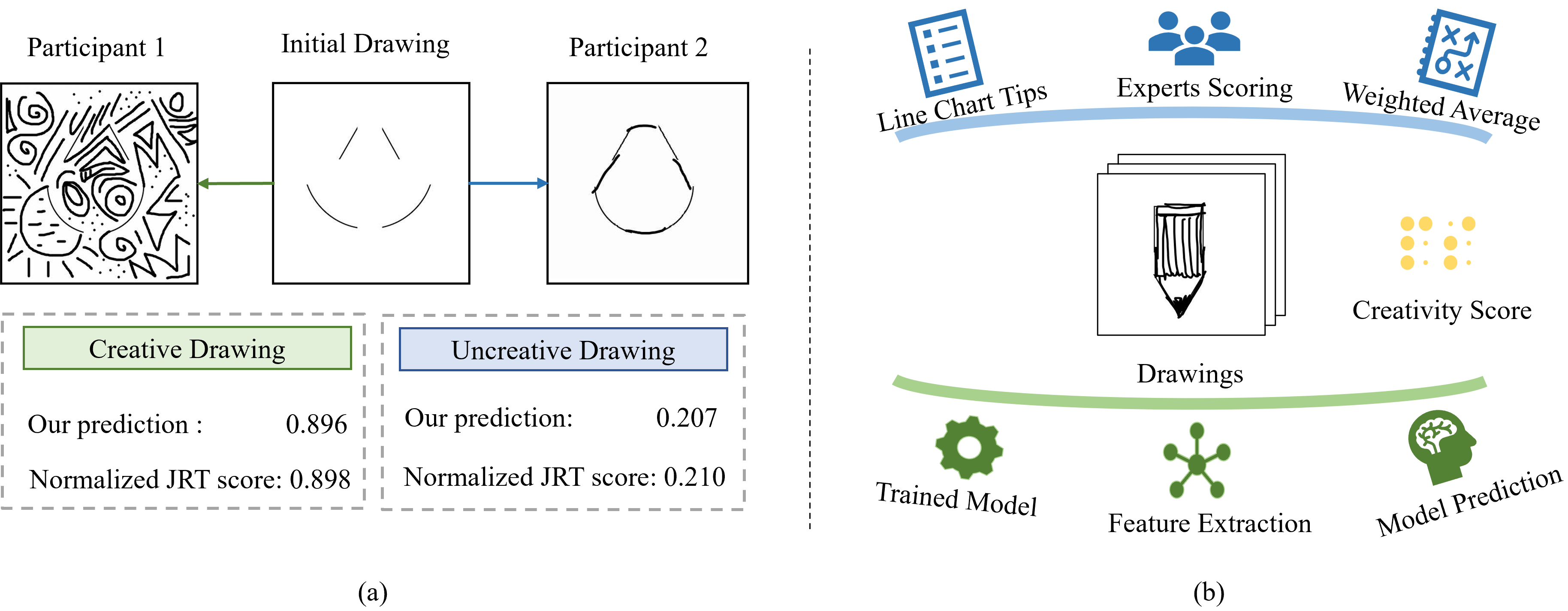}
  \vspace{-1.5em}
  \caption{(a) Example drawings from a figural creativity task rated as creative (left) versus uncreative (right). (b) Comparison of evaluation processes: the traditional expert-based scoring pipeline (top) versus our proposed automated assessment model (bottom). }
  \label{fig:concept}
\end{figure}

However, this manual scoring process has significant drawbacks. Expert evaluation of drawings is time-consuming, costly, and prone to inconsistency~\cite{patterson2024audra,baer_consistency}. Each drawing must be examined by trained raters who apply complex rubrics (e.g., for original form, narrative content, or expressive style)~\cite{baer_consistency}. Inter-rater agreement can be low without extensive training, and scaling to large datasets is infeasible in practice~\cite{patterson2024audra,baer_consistency}. As illustrated in Fig.~\ref{fig:concept}, the traditional pipeline (Fig.~\ref{fig:concept}a) contrasts sharply with our proposed automated approach (Fig.~\ref{fig:concept}b): human experts must annotate each drawing one by one, whereas a machine learning model can process thousands of images quickly and uniformly.

To address these issues, recent studies have begun exploring automated creativity scoring using machine learning~\cite{patterson2024audra,panfilova_resnet,zhang_painting,cropley_figural}. For example, Patterson et al. trained a convolutional neural network (AuDrA) on over 13{,}000 line drawings with human ratings, achieving high correlation with expert scores on new sketches~\cite{patterson2024audra}. Panfilova et al. fine-tuned deep networks (e.g., ResNet, MobileNet) on drawings from the “Urban” creative test and applied explainable AI techniques (Grad-CAM) to interpret model focus~\cite{panfilova_resnet}. Zhang et al. developed a CNN to assess the creativity of paintings, achieving 90\% accuracy on a small set of professional and child artworks~\cite{zhang_painting}. Similarly, Cropley et al. constructed a large image-based classifier for the Figural Test of Creative Thinking, reporting performance that exceeds average human agreement~\cite{cropley_figural}.

Despite these advances, existing automated models have notable limitations. Many rely on relatively shallow architectures or simple classification schemes~\cite{zhang_painting,cropley_figural}, making it hard to capture the full nuance of creative expression. In particular, black-box CNN scores are not readily interpretable~\cite{panfilova_resnet,chen_bias}, and most models are trained on one specific task or dataset, raising concerns about generalization~\cite{patterson2024audra}. Models may inadvertently learn spurious cues (e.g., ink coverage) that do not reflect true creativity, leading to poor transfer across tasks~\cite{elgammal_art}. This can also be verified by \cite{Pencils_to_Pixels}, which shows that automated tools overfit on ink density. In summary, we observe three key gaps: shallow modeling, lack of interpretability, and limited cross-task robustness~\cite{gan_generalization}.

Nath et al.~\cite{Pencils_to_Pixels} present a key content–style framework for assessing creativity in drawings. They measure four stylistic attributes—ink density, fraction of ink within the stimulus shape, number of connected components, and number of lines extracted via skeletonization and Hough transforms. For content, they use CLIP image embeddings to capture semantic similarity to the base stimulus and generate short GPT-4o captions. These interpretable content and style descriptors are then combined in a regression model to predict human creativity ratings.

%This view posits that creativity in a drawing arises from both \textit{what} is depicted (content) and \textit{how} it is depicted (style). For example, two sketches of a forest may contain the same content (trees and animals), but a richly textured or unusual style can make one appear more creative than the other. These observations align with findings in computational creativity and vision (e.g., neural style transfer separates content and style~\cite{gatys_styletransfer}). Therefore, instead of treating creativity as a monolithic score, we explicitly factor our model to capture content semantics (objects) and stylistic features (line complexity).

Building on~\cite{Pencils_to_Pixels}, we design a multi-task learning architecture. One head of the network predicts the overall creativity score, another predicts the drawing’s content category (e.g., ``plant,” ``animal”, ``object”), and another extracts quantifiable style descriptors (line complexity). The use of multi-task learning allows shared representations to benefit all predictions, while each task provides an interpretability signal. Crucially, we incorporate a conditional learning mechanism that dynamically modulates the image embedding based on identified content and style cues. {Our main contributions are:}
\begin{itemize}[label=\textbullet]
    \item  We systematically augment the existing drawing image dataset with semantic content categories labeled to guide the model in learning creativity from the content perspective. \vspace{0.5\baselineskip}
    \item We introduce a unified, multi-task model for visual creativity assessment that predicts the content and style components of drawings simultaneously.  \vspace{0.5\baselineskip}
    \item We develop a conditional multi-task learning architecture that adaptively focuses on creativity-relevant signals depending on content and style cues. \vspace{0.5\baselineskip} 
    \item We demonstrate through experiments that our approach outperforms prior methods in both scoring accuracy and cross-dataset generalization.
\end{itemize}

\section{Method}

\subsection{Dataset Augmentation}
The publicly available dataset AuDrA-Drawings dataset \cite{patterson2024audra} is adopted for our model effectiveness verification. However, the content and style annotations in this dataset are missing.
To guide the model in learning visual features from two orthogonal dimensions—content and style—we established a labeling team consisting of six systematically trained undergraduate students.  Each image is annotated with five mutually exclusive content labels ( ``human", ``plant", ``animal", ``object", and ``other") based on first-impression visual content. All annotators adhered to standardized procedures to ensure annotation regularity and reliability. For style, we empirically quantified the ink intensity on the canvas, under the mild assumption that the drawing style is correlated with the number and complexity of strokes. This scalar value is treated as a proxy that encodes style-related information.
\begin{figure}[htp]
\centering
\includegraphics[width=\linewidth]{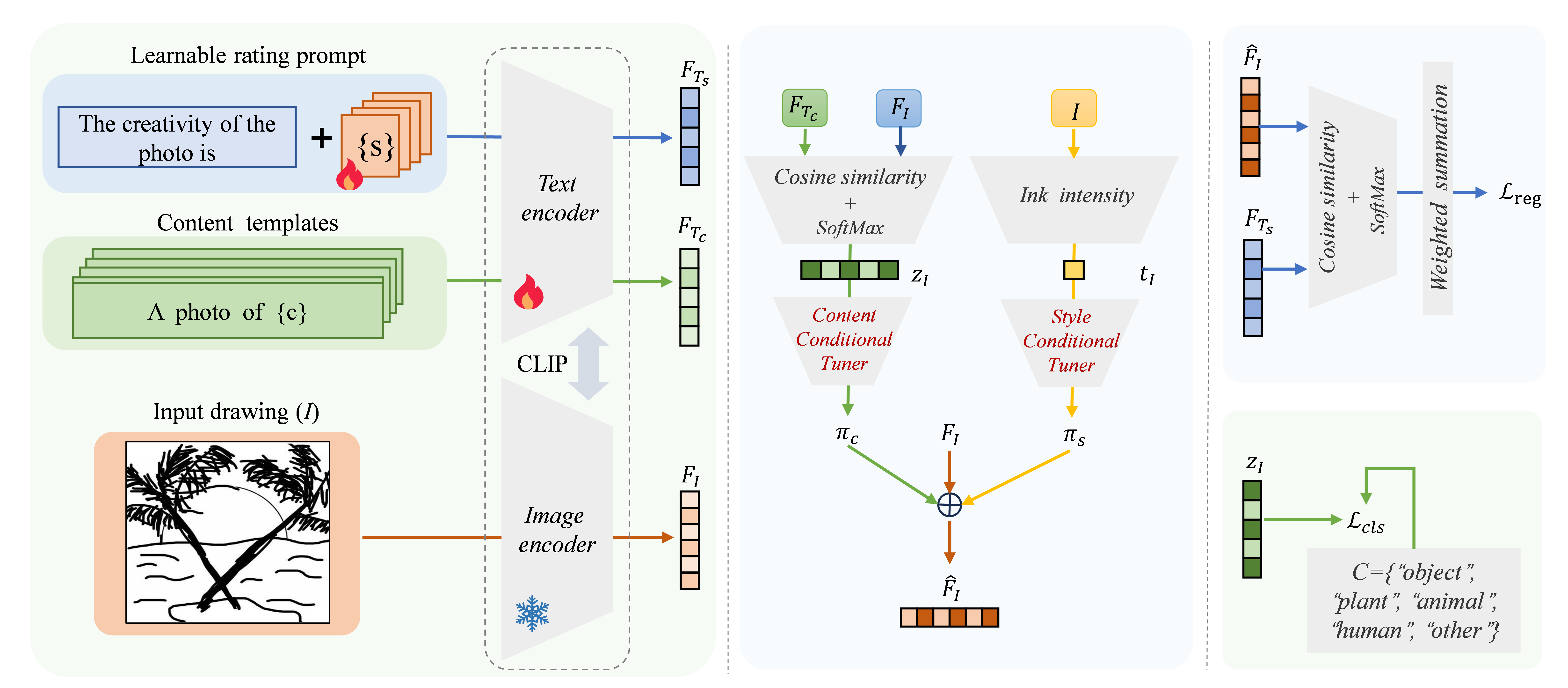}
\caption{Framework of our proposed \textit{Content-Style conditioned Creativity Assessment} (CSCA) model. The model fuses CLIP-based vision and text embeddings with content and style-conditioned tuning modules to predict creativity scores in a cross-modal, interpretable manner.}
\label{fig:architecture}
\end{figure}

\subsection{Overview of Framework}
The overall architecture of our \textit{Content-Style conditioned Creativity Assessment} (CSCA) model is illustrated in Fig.~\ref{fig:architecture}. Given an input drawing image $I \in \mathbb{R}^{ H \times W\times3}$, our model aims to predict a creativity score $\hat{q}(I) \in [0, 1]$, trained to approximate the normalized ground-truth creativity score $q(I)$ derived from expert judgments. Our method builds upon the CLIP framework, which consists of an image encoder $E_I(\cdot)$ and a text encoder $E_T(\cdot)$. To enhance both interpretability and prediction accuracy, we design a multi-branch architecture that integrates semantic priors from two orthogonal dimensions: \emph{content} and \emph{style}. Specifically, we first define two textual  templates to prompt the multi-task learning :
\begin{itemize}
    \item \textbf{Content prompt:} \texttt{"a photo of \{c\}"} for each $c \in C = \{``\text{object}",``\text{plant}",  \\ ``\text{animal}",``\text{human}", ``\text{other}", \}$, where $C$ denotes the predefined set of content categories.
    \item \textbf{Creativity score prompt:} \texttt{"the creativity of the photo is \{s\}"} for each $s$ is initialized from $S = \{``\texttt{bad}", ``\texttt{poor}", ``\texttt{fair}", ``\texttt{good}", ``\texttt{perfect}"\}$, corresponding to  creativity score weights $w_s \in W_s = \{0.2, 0.4, 0.6, 0.8, 1.0\}$.
\end{itemize}

To move beyond handcrafted prompts and enhance adaptability, we propose the following three main components:
\begin{itemize}
    \item \textbf{Learnable creativity rating embeddings:} A set of learnable text tokens $\{\delta_s\}_{s \in S}$ that replace fixed templates in $S$ with semantically-rich embeddings aligned with visual concepts via $E_T(\cdot)$.
    \item \textbf{Content Conditional Tuner:} A lightweight network that generates a modulation vector $\pi_c \in \mathbb{R}^d$ based on the predicted content category $c$, where $d$ is the embedding dimension.
    \item \textbf{Style Conditional Tuner:} Another modulation module that outputs a vector $\pi_s \in \mathbb{R}^d$ conditioned on a scalar style proxy $t_I$, which reflects the density of ink-based stroke elaboration.
\end{itemize}
The final modulated visual embedding $\hat{F}_I \in \mathbb{R}^d$ is tuned by:
\begin{equation}
\hat{F}_I = {F}_I  + \pi_c + \pi_s,
\end{equation}
with 
\begin{equation}
{F}_I =  E_I(I),
\end{equation}
where $E_I(I)$ is the visual representation from the image encoder, and $\pi_c$, $\pi_s$ are the content and style modulation vectors, respectively.

To assess creativity, we compute the similarity between $\hat{F}_I$ and each creativity embedding $F_{T_s}$($F_{T_s} = E_T(\delta_s)$) using cosine similarity, followed by softmax normalization:
\begin{equation}
{p}(s\,|\,x) = \frac{\exp\left( \mathrm{sim}(F_I, F_{T_s}) / \tau \right)}{\sum\limits_{s=1}^{|S|} \exp\left( \mathrm{sim}(F_I, F_{T_s}) / \tau \right)},
\label{eq:softmax_prob}
\end{equation}
where $\tau$ is a temperature scaling factor and $\mathrm{sim}(\cdot,\cdot)$ denotes cosine similarity. The final creativity score $\hat{q}(I)$ is computed as the expected value over the predicted distribution:
\begin{equation}
\hat{q}(I) = \sum_{s \in S} \hat{p}(s\,|\,I) \cdot w_s.
\label{eq:score_estimate}
\end{equation}
The details of each module are described as follows.\\

\subsection{Design Details}
\noindent \textbf{(1) Learnable Creativity Rating Embedding.}
Manual prompt engineering often fails to deliver precise semantic disentanglement across tasks. For example, the term good might ambiguously signal either creative quality or technical skill. To overcome this limitation, we propose learnable creativity rating embeddings. Specifically, five creativity-level phrases (``bad", ``poor", ``fair", ``good", ``perfect") are treated as trainable embedding vectors $\{\delta_s\}_{s \in S}$ rather than static templates. These vectors are optimized jointly with the rest of the model, allowing for cross-modal alignment and richer semantic grounding. This design allows the model to automatically learn the semantics of creativity in an end-to-end fashion, enhancing the quality and interpretability of creativity predictions.\\

\noindent \textbf{(2) Content Conditional Tuner.} We propose a content-conditional tuner to inject semantic priors from the image’s content into the visual embedding space. For each input image $I$, we compute the cosine similarity between its visual embedding ${F}_I$ and a set of text embeddings ${F_{T_c}}$ (c=1,2,.., $\left|C\right|$) derived from the content prompts.
\begin{equation}
logit(c|\mathbf{x}) = \frac{\exp(\mathrm{sim}({F}_I, F_{T_c})/\tau)}{\sum_{c=1}^{|C|} \exp(\mathrm{sim}(F_{I}, F_{T_c})/\tau)}.
\label{logit_c}
\end{equation}
Let ${z_I} \in \mathbb{R}^{|C|}$ denote the softmax results across content categories of image $I$. This vector is passed through a two-layer content meta-network $f_c$ to generate the conditional embedding $\pi_c$:
\begin{equation}
\pi_c = f_c({z_I}).
\end{equation}
This content-aware embedding biases the visual feature $F_I$ to be more sensitive to content-relevant cues (e.g., human figure composition vs. abstract object layout), offering interpretable and task-relevant modulation.\\

\noindent \textbf{(3) Style Conditional Tuner.} 
Inspired by the finding that style (i.e., elaboration or ink quantity) correlates with creativity \cite{patterson2024audra}, we further propose a style-conditional tuner for the visual embedding. Let $t_I$ be a scalar representing the style value of image $I$.  This scalar is further passed through a style meta-network, 
%which is implemented as a lightweight two-layer multilayer perceptron (MLP) followed by a ReLU activation layer and produces a style-conditioned embedding $\pi_s$:
which is implemented as a lightweight structure that first applies a sigmoid layer, followed by a fully-connected (FC) layer, a ReLU activation layer, and another FC layer, producing a style-conditioned embedding $\pi_s$:
\begin{equation}
\pi_s = f_s(t_I).
\end{equation}
This embedding supplements the visual representation with low-level stroke density, which CLIP may otherwise ignore. Fusing $\pi_s$ with $F_I$ facilitates richer representation across style-content axes.

\subsection{Loss for Multitask Learning} 
The training objective is a combination of score regression loss and content classification loss. Given a batch of $N$ drawings $\mathcal{I} = {\{I_1, I_2, ..., I_N\}}$ with true creativity scores $q(I) = \{q(I_1),q(I_2), ..., q(I_N)\}$ and true content labels $l(I) = \{l(I_1), l(I_2), ..., l(I_N)\}$, we define:

\noindent Creativity regression loss:
\begin{equation}
\mathcal{L}_{\text{reg}} = \frac{1}{N} \sum_{i=1}^{N} \left( q(I_i) - \hat{q}(I_i) \right)^2.
\end{equation}
Content classification loss:
\begin{equation}
\mathcal{L}_{\text{cls}} = -\frac{1}{N} \sum_{i=1}^{N} \sum_{c=1}^{|C|} \mathbf{1} (l(I_i) = c) \cdot \log \left( ({z}_{(I_i)})_c \right),
\end{equation}
where $({z}_{(I_i)})_c $ denote the probability that  image $I_i$ categorized into the $c$-th content.  The final multitask loss is defined by:
\begin{equation}
\mathcal{L} =  \mathcal{L}_{\text{reg}} +  \lambda \cdot \mathcal{L}_{cls},
\label{all}
\end{equation}
where $\lambda \in [0,1]$ balances the regression and classification objectives.

\section{Experiments}

\subsection{Experimental Settings}

In this study, we employed the AuDrA-Drawings dataset \cite{patterson2024audra}, which comprises over 13,000 drawing–rating pairs across four subsets. The \textbf{Primary Dataset} serves as the core, containing 11,075 abstract drawings. It was split into training (70\%), validation (10\%), and test (20\%) sets following standard machine learning protocols. We used the training set for model learning, the validation set for early stopping and hyperparameter tuning, and the test set for final evaluation. Each drawing was rated for creativity on a 1–5 scale by 50 rigorously trained undergraduate students, achieving high inter-rater reliability (Intraclass Correlation Coefficient (ICC) $>$ 0.89).

The remaining three subsets are designed to evaluate different aspects of model generalization. The \textbf{Rater Generalization 1 (RG1)} subset includes 670 abstract drawings rated by 3 new raters (ICC = 0.73), assessing model performance on unseen raters. The \textbf{Rater Generalization 2 (RG2)} subset contains 722 abstract drawings rated by 6 new raters (ICC = 0.90), further testing rater-based generalization. The \textbf{Rater \& Task Generalization (FG)} subset consists of 679 drawings of specific objects (rather than abstract ones), rated by 3 new raters (ICC = 0.63), allowing joint evaluation of generalization to new raters and new task types. The summary of the subsets is shown in Table~\ref{tab:1}.

\begin{table}
\centering
\caption{Summary of AuDrA-Drawings dataset.}
\begin{tabular}{lcccc}
\toprule
\textbf{Subset} & \textbf{Drawing Type} & \textbf{\# Samples} & \textbf{Raters} & \textbf{ICC} \\
\midrule
Primary Dataset & Abstract         & 11,075 & 50 trained undergraduates & $>$ 0.89 \\
Rater Gen. 1    & Abstract         & 670    & 3 new raters              & 0.73 \\
Rater Gen. 2    & Abstract         & 722    & 6 new raters              & 0.90 \\
Rater \& Task Gen. & Specific Objects & 679    & 3 new raters              & 0.63 \\
\bottomrule
\label{tab:1}
\end{tabular}
\end{table}

For a fair comparison, we maintain the training settings consistent with the AuDrA benchmark: a batch size of 16, a learning rate of 1e-5, and a maximum of 136 epochs. We set $\lambda$ by 1e-3 in Eqn.~(\ref{all}). We adopted ViT-L/14 as the visual encoder and kept it frozen throughout training. Input drawings were preprocessed by inverting colors (black ink rendered as high pixel values and white backgrounds as low), standardizing RGB channels of the generalization datasets based on training set statistics to ensure consistent global complexity, and resizing all images to $224\times224$. Model performance was measured using Spearman Rank Correlation Coefficient (SRCC) and Pearson Linear Correlation Coefficient (PLCC) between predicted creativity scores and normalized Joint Rater Truth (JRT) ratings. Let the predicted scores be $\hat{Q} = \begin{pmatrix} \hat{q}(x_1) & \hat{q}(x_2) & \cdots & \hat{q}(x_n) \end{pmatrix}$ and the corresponding normalized JRT scores be  $Q = \begin{pmatrix} q(x_1) & q(x_2) & \cdots & q(x_n) \end{pmatrix}.$ Then, SRCC and PLCC are defined as:
\begin{equation}
 \quad S R C C=\frac{\sum_i^n\left(r_i-\bar{r}_{i}\right)\left({s_i}-\bar{s}_{i}\right)}{\sqrt{\sum_{i=1}^n\left(r_i-\bar{r}_{i}\right)^2} \sqrt{\sum_{i=1}^n\left(s_i-\bar{s}_{i}\right)^2}},
\end{equation}

\begin{equation}
P L C C = \frac{\sum_{i=1}^{n} (\hat{q}_i - \bar{\hat{q}})(q_i - \bar{q})}{\sqrt{\sum_{i=1}^{n} (\hat{q}_i - \bar{\hat{q}})^2} \sqrt{\sum_{i=1}^{n} (q_i - \bar{q})^2}},
\end{equation}
where $r_i$ and $s_i$ are the ranks of $\hat{Q}$ and $Q$, respectively.

\subsection{Experimental results}

\subsubsection{Performance Comparison on Primary Set.}
To rigorously assess the effectiveness of our proposed method (CSCA), we conducted a comprehensive evaluation on the primary test set, where multiple baseline models across different architectures were reimplemented for direct comparison under consistent training settings. Specifically, we employed five CLIP-based architectures (ViT-B/32, ViT-B/16, ViT-L/14, RN50, and RN101) following the CLIPIQA+ strategy \cite{wang2023exploring}, which maps image embeddings to predefined text prompts (``a photo of good creativity" and ``a photo of bad creativity") and computes the similarity score via cosine similarity and softmax normalization. In addition, we included recent state-of-the-art creativity assessment and image quality assessment models such as Audra \cite{patterson2024audra} and AGIQA \cite{tang2025clip} as reference points. All models were optimized using MSE loss to minimize the prediction error with respect to ground-truth creativity annotations.

As reported in Table \ref{tab2}, our method achieves an SRCC of 0.86 and PLCC of 0.87, surpassing all CLIP-based variants (maximum SRCC and PLCC of 0.81 from CLIP-RN101) and even outperforming the recent Audra model (SRCC/PLCC = 0.80/0.79). This consistent improvement across both correlation metrics strongly evidences the superiority of our content-style conditioned creativity assessment framework. 
%The results validate that CACA can more effectively align visual representations with nuanced semantic cues necessary for evaluating creativity in images.
\begin{table}[h]
    \centering
    \caption{Performance compression on the primary test set.}
    \label{tab2}
    \begin{tabular}{c|c|c|c|c}
        \hline
        \textbf{Group} & \textbf{Model} & \textbf{Year} & \textbf{SRCC} & \textbf{PLCC} \\
        \hline
        \multirow{1}{*}{\makecell[c]{VGG \cite{simonyan2014very} }}
         &VGG16   & 2014 &  0.52 & 0.04  \\
        \hline
        \multirow{2}{*}{\makecell[c]{ResNet \cite{he2016deep}}}
        & ResNet34       & 2016 &  0.79 & 0.81  \\
        & ResNet50       & 2016 &  0.78 & 0.79  \\
        \hline
        \multirow{5}{*}{\makecell[c]{{CLIPIQA+}} \cite{patterson2024audra}} 
        & CLIP(ViT-B/32) & 2023 & 0.63 & 0.65 \\
        & CLIP(ViT-B/16) & 2023 & 0.67 & 0.68 \\
        & CLIP(ViT-L/14) & 2023 & 0.67 & 0.69 \\
        & CLIP(RN50)     & 2023 & 0.80 & 0.81 \\
        & CLIP(RN101)    & 2023 & 0.81 & 0.81 \\
        \hline
        \multirow{2}{*}{\makecell[c]{{Others}}} 
        & Audra \cite{patterson2024audra}         & 2024 & 0.80 & 0.79 \\
        & AGIQA \cite{tang2025clip}         & 2025 & 0.79  &0.80   \\
        &Pencils to Pixels \cite{Pencils_to_Pixels}  & 2025 & 0.79  &-\\
        \hline
        \makecell[c]{\textbf{Ours}} 
        & CSCA   & 2025 & \textbf{0.86} & \textbf{0.87} \\
        \hline
    \end{tabular}
\end{table}

\subsubsection{Generalization on the Other Three Sets.}
\begin{table}[h]
    \centering
    \caption{Generalization capability on the other three subsets in terms of SRCC and PLCC.}
    \label{tab3}
    \begin{tabular}{c|c|c|cc|cc|cc}
        \hline
        \textbf{Group} & \textbf{Model} & \textbf{Year} 
        & \multicolumn{2}{c|}{\textbf{RG1}} 
        & \multicolumn{2}{c|}{\textbf{RG2}} 
        & \multicolumn{2}{c}{\textbf{FG}} \\
        \cline{4-9}
        & & & SRCC & PLCC & SRCC & PLCC & SRCC & PLCC \\
        \hline
        \multirow{1}{*}{\makecell[c]{VGG\cite{simonyan2014very}}}
         & VGG16   & 2014 & 0.58 & 0.56 & 0.39 & 0.32 & 0.21 & 0.12 \\
        \hline
        \multirow{2}{*}{\makecell[c]{ResNet\cite{he2016deep}}}
        & ResNet34       & 2016 & 0.78 & 0.75 & 0.68 & 0.66 & 0.45 & 0.47 \\
        & ResNet50       & 2016 & 0.80 & 0.77 & 0.67 & 0.65 & 0.46 & 0.47 \\
        \hline
        \multirow{5}{*}{\makecell[c]{CLIPIQA+\cite{patterson2024audra}}}
        & CLIP(ViT-B/32) & 2023 & 0.69 & 0.65 & 0.41 & 0.43 & 0.43 & 0.44 \\
        & CLIP(ViT-B/16) & 2023 & 0.73 & 0.70 & 0.46 & 0.47 & 0.37 & 0.39 \\
        & CLIP(ViT-L/14) & 2023 & 0.72 & 0.69 & 0.47 & 0.48 & 0.35 & 0.36 \\
        & CLIP(RN50)     & 2023 & 0.77 & 0.75 & 0.67 & 0.65 & 0.43 & 0.45 \\
        & CLIP(RN101)    & 2023 & 0.79 & 0.76 & 0.68 & 0.66 & 0.37 & 0.38 \\
        \hline
        \multirow{2}{*}{\makecell[c]{Others}} 
        & Audra \cite{patterson2024audra} & 2024 & 0.78 & 0.75 & 0.68 & 0.67 & 0.44 & 0.46 \\
        & AGIQA \cite{tang2025clip}       & 2025 & 0.78 & 0.77 & 0.68 & 0.67 & 0.44 & 0.45 \\
        \hline
        \makecell[c]{\textbf{Ours}} 
        & CSCA & 2025 & \textbf{0.82} & \textbf{0.79} & \textbf{0.74} & \textbf{0.73} & \textbf{0.48} & \textbf{0.49} \\
        \hline
    \end{tabular}
\end{table}
Beyond primary test set performance, we further examined the generalization ability of our model under two key scenarios: (1) new-sample-same-task-new-rater (RG1 and RG2 datasets) and (2) new-sample-cross-task-new-rater (FG dataset). These test conditions evaluate whether a model can maintain performance when confronted with unseen images rated by different annotators or requiring generalization across different creativity evaluation tasks.

As summarized in Table \ref{tab3}, our proposed model achieves superior results on all three generalization datasets, with SRCC/PLCC scores of 0.82/0.79 on RG1, 0.74/0.73 on RG2, and 0.48/0.49 on FG. Notably, on the RG2 and FG sets—where semantic shifts and rater variance are most pronounced—our model exhibits substantial performance gains over all CLIPIQA+ baselines and even marginally outperforms the Audra model. For example, on the FG dataset, CSCA achieves a PLCC of 0.49 compared to Audra (0.46) and CLIP-RN101 (0.38). These results highlight the strong generalization capacity of our approach across rater distributions and task variations, underscoring its robustness in practical deployment scenarios.

\subsection{Ablation Studies}

To rigorously validate the  soundness of our proposed model architecture, we conducted a comprehensive ablation study. Starting from the baseline CLIP model (ViT-L/14), we progressively enhanced the architecture through a staged integration of key components: (a) Incorporation of a Learnable Creativity Rating embedding (LCR) to establish a semantic correlation layer addressing subjective rater bias;  
(b) Addition of a Style Conditional Tuner (SCT) aimed at capturing physical attributes such as ink quantity;  
(c) Introduction of a Content Conditional Tuner (CCT) designed to extract content-aware semantic features;  
(d) The final model jointly integrates both SCT and CCT modules, constituting the full proposed architecture. This structured approach allows us to isolate and assess the individual and combined impact of each module on visual creativity evaluation.

\begin{table}[htp]
    \centering
    \caption{PLCC results of the ablation study. \textbf{LCR}: Learnable Creativity Rating embedding, \textbf{SCT}: Style Conditional Tuner, \textbf{CCT}: Content Conditional Tuner. Only the best score per column is bolded.}
    \label{tab4}
 
    \rowcolors{2}{gray!10}{white}
    \resizebox{0.9\textwidth}{!}{
    \begin{tabular}{l|c|c|c|c|c|c|c}
        \hline
        {Model ID} & {LCR} & {SCT} & {CCT} & {Primary} & {RG1} & {RG2} & {FG} \\
        \hline
        (1) CLIP Baseline      & \ding{55} & \ding{55} & \ding{55} & 0.69  & 0.69 & 0.48 & 0.36 \\
        (2) w/ LCR       & \ding{51} & \ding{55} & \ding{55} & 0.86  & 0.77 & 0.71 & 0.47 \\
        (3) w/ LCR + SCT       & \ding{51} & \ding{51} & \ding{55} & 0.86  & 0.78 & 0.72 & \textbf{0.51} \\
        (4) w/ LCR + CCT       & \ding{51} & \ding{55} & \ding{51} & 0.86  & 0.76 & 0.72 & 0.47  \\
        (5) Our approach       & \ding{51} & \ding{51} & \ding{51} & \textbf{0.87}  & \textbf{0.79} & \textbf{0.73} & 0.49 \\
        \hline
    \end{tabular}
    }
\end{table}

Table~\ref{tab4} presents the PLCC results across the four subsets. The baseline CLIP model (1) attains a PLCC of 0.69 on the Primary dataset but experiences substantial performance degradation on generalization datasets, with a notably low score of 0.36 on the FG dataset. This clearly reveals its significant limitations in handling rater subjectivity and task variability.
The model (2) with the sole introduction of the LCR module significantly improves the PLCC to 0.86 on the Primary dataset and 0.77 on the RG1 dataset. This indicates that explicitly modeling creativity ratings can remarkably enhance the semantic alignment between the model and rater judgments. However, its score of 0.47 on the FG dataset suggests persistent challenges in cross-task generalization.
The model (3) integrating the SCT module with LCR further boosts the performance on the FG dataset to 0.51. This is attributed to the SCT module's ability to effectively model physical style cues, such as ink density, thereby enhancing the model's robustness against rater differences within the same task. The model (4) using the CCT module independently improves the performance on the RG2 dataset to 0.72, highlighting the importance of content-aware semantic features in adapting to task shifts.
Finally, the full model (5) combining both SCT and CCT achieves the highest PLCC of 0.87 on the Primary dataset and maintains consistently strong performance across all datasets. This demonstrates that the complementary integration of semantic content and physical style modeling captures both high-level creativity concepts (e.g., narrative structure in abstract shapes) and low-level visual cues (e.g., line density), significantly enhancing the model's capabilities in cross-rater consistency and cross-task generalization.

\subsection{Visualization}

\subsubsection{Style and Creativity Correlation}

\begin{figure}[htp]
\centering
\includegraphics[width=\linewidth]{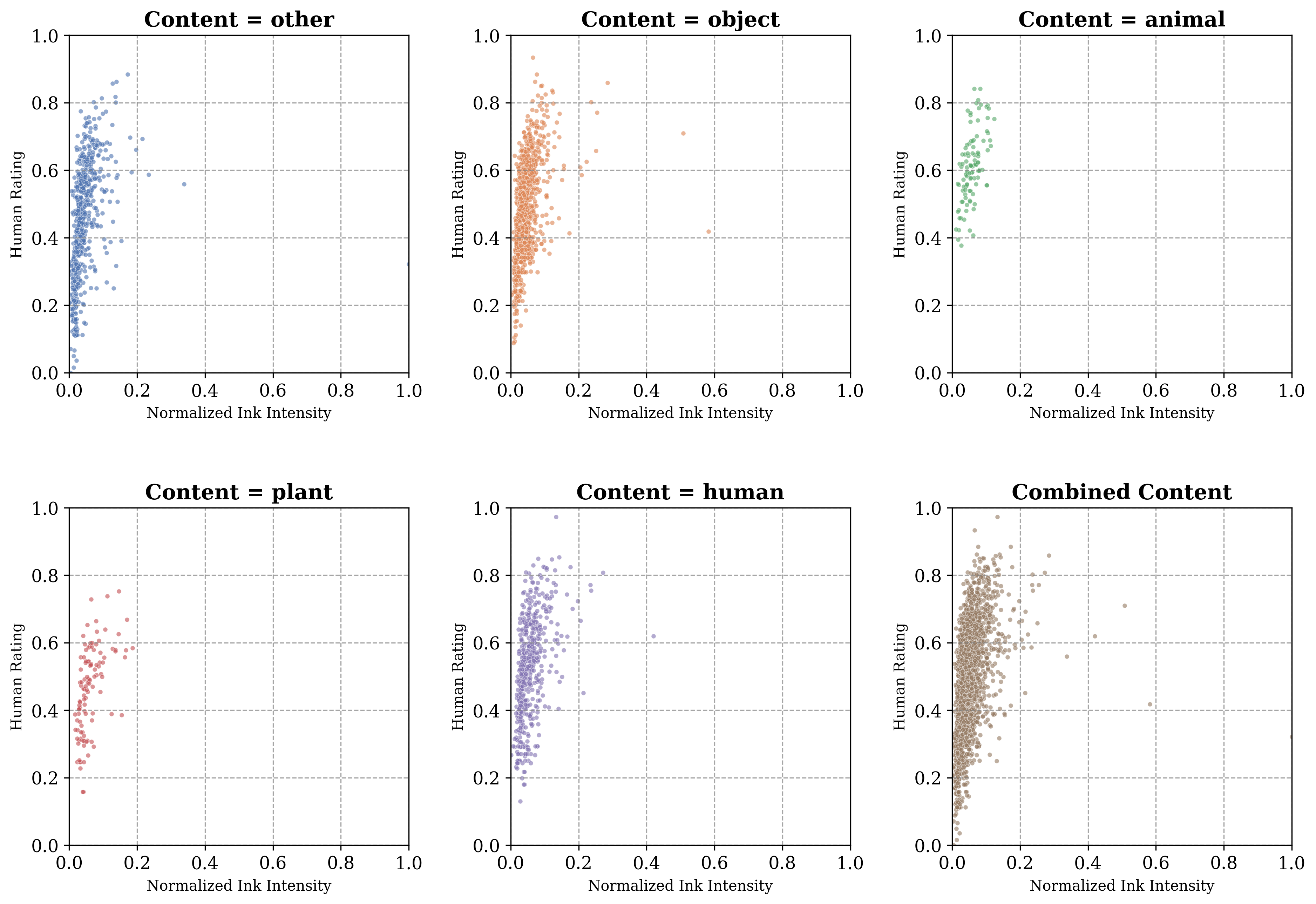}
\caption{Correlation between normalized ink intensity and human creativity ratings across different content categories.}
\label{fig:ink_rating}
\end{figure}

To better understand the relationship between visual style and creativity perception, we analyzed the test set of the Primary Dataset by examining the correlation between normalized ink intensity and human-assigned creativity scores. Fig.~\ref{fig:ink_rating} presents scatter plots across five content categories (`other', `object', `animal', `plant', and `human'), as well as the overall data distribution.

We also computed Spearman's rank correlation coefficient (SRCC) and associated p-values for each content category and for the entire dataset. As shown in Table~\ref{tab:srcc_ink}, SRCC values ranged from 0.54 to 0.66 across individual content categories, with an overall SRCC of 0.60. All correlations were statistically significant ($p \ll 0.05$), indicating a strong and consistent positive association between ink intensity and human creativity ratings. These findings reinforce the idea that visual density plays a meaningful role in perceived creativity. These results suggest that ink intensity is a reliable low-level visual cue correlated with perceived creativity across diverse content types, supporting its integration as a key feature in automated creativity assessment models.

\begin{figure}[htp]
\centering
\includegraphics[width=0.7\linewidth]{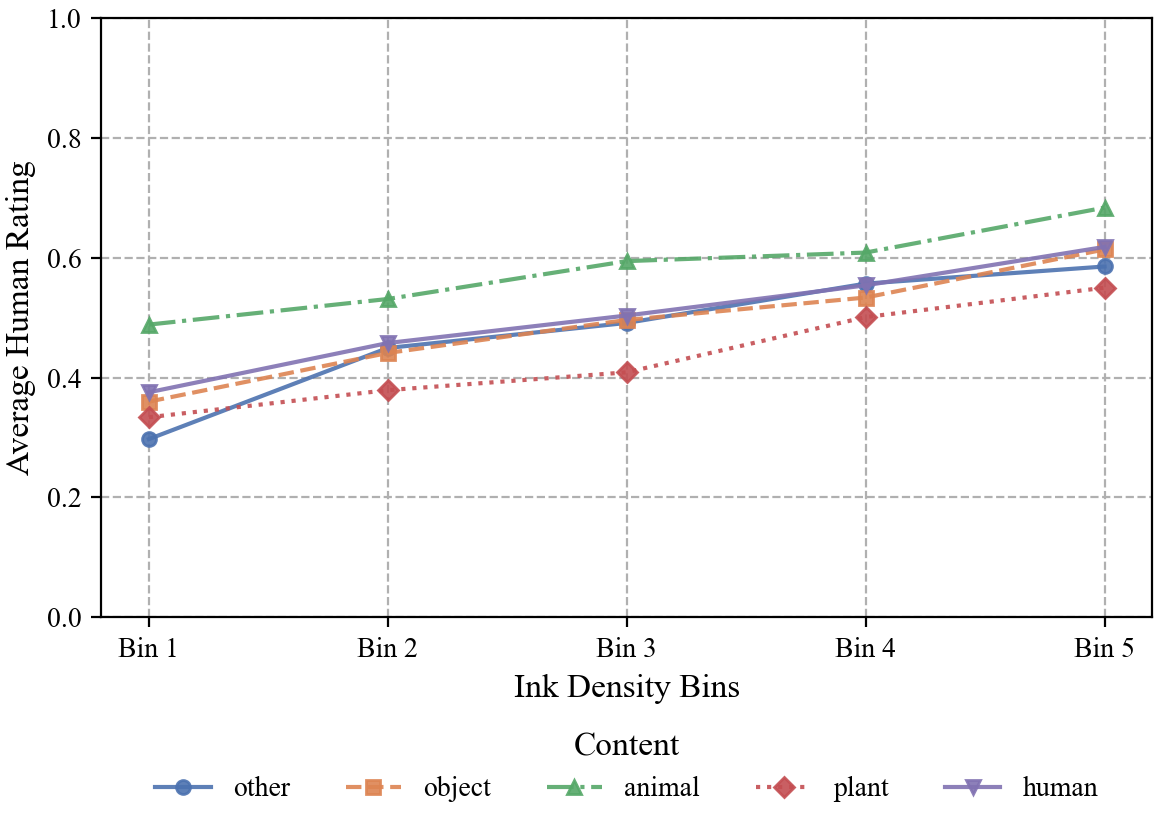}
\caption{Mean human creativity ratings across content categories, grouped by ink intensity levels.}
\label{fig:avg_human_rating}
\end{figure}

\begin{table}[htp]
\centering
\caption{Spearman's rank correlation coefficients (SRCC) between normalized ink intensity and human creativity ratings across content categories.}
\label{tab:srcc_ink}
\begin{tabular}{lccc}
\toprule
\textbf{Content Category} & \textbf{SRCC} & \textbf{p-value} \\
\midrule
Other   & 0.66 & $1.30 \times 10^{-72}$ \\
Object  & 0.58 & $3.44 \times 10^{-83}$ \\
Animal  & 0.59 & $4.15 \times 10^{-11}$ \\
Plant   & 0.63 & $1.98 \times 10^{-13}$ \\
Human   & 0.54 & $6.28 \times 10^{-40}$ \\
\midrule
\textbf{Combined} & \textbf{0.60} & $\mathbf{2.03 \times 10^{-218}}$ \\
\bottomrule
\end{tabular}
\end{table}

\subsubsection{Content-Type and Rating Trends}

To further investigate how content interacts with creativity perception, we grouped the test-set drawings into five bins based on ink intensity. Within each bin, samples were further divided by content type. We then computed the average human creativity rating for each combination. The results are visualized in Fig.~\ref{fig:avg_human_rating}.

Overall, creativity ratings increase with ink intensity across all content types, consistent with the earlier correlation analysis. Notably, drawings labeled as `animal' consistently receive higher average creativity scores compared to other categories. This may reflect a richer potential for variation in animal depictions (e.g., poses, anatomy), which could foster stronger perceptions of originality. In contrast, the `plant' and `other' categories tend to yield lower creativity ratings, possibly due to more constrained forms or less coherent visual themes.

\section{Conclusion}

In this paper, we present a cognitively inspired framework for automatic creativity assessment in human drawings, integrating both \textit{what} is drawn (content) and \textit{how} it is drawn (style). By enriching an existing dataset with semantic annotations and introducing a conditional multi-task architecture, our model jointly predicts creativity scores while attending to both stylistic and conceptual cues. Ablation studies confirm the effectiveness of each proposed module—LCR, CCT, and SCT—in enhancing performance and generalization across raters and tasks. The full model outperforms baselines and provides interpretable attention aligned with human intuition. Our work offers a scalable and explainable alternative to subjective expert scoring. 
Future research could explore cross-cultural generalization of creativity judgments, personalization to specific rater profiles, and extending the framework to multi-modal creative domains such as storytelling or musical composition. We believe this work offers a concrete step toward human-centered, cognitively-informed AI systems for creativity evaluation.

\subsubsection{Revision Note}
This version includes expanded related work and clarifications acknowledging the contributions of Nath et al.~\cite{Pencils_to_Pixels}. The PRCV 2025 conference version was prepared earlier and does not contain these updates; the arXiv version should be considered the most complete and up-to-date reference.

\subsubsection{Acknowledgments}This work was supported in part by the National Natural Science Foundation of China under Grant 62401214,  in part by grants from Project of Key Institute of Humanities and Social Science, MOE (16JJD880025).

\bibliographystyle{splncs04}
\bibliography{new_ref}
\end{document}